\documentclass[journal]{IEEEtran}
\usepackage{amsmath,amsfonts}
\usepackage{algorithmic}
\usepackage{algorithm}
\usepackage{array}
\usepackage{textcomp}
\usepackage{stfloats}
\usepackage{url}
\usepackage{verbatim}
\usepackage{graphicx}
\usepackage{cite}
\usepackage{multirow}
\usepackage{booktabs}
\usepackage{amssymb}
\usepackage{subfigure}
\usepackage{flushend}
\usepackage{color}

\begin{document}

\title{GBE-MLZSL: A Group Bi-Enhancement\\
 Framework for Multi-Label Zero-Shot Learning}

\author{
Ziming Liu,~\IEEEmembership{Student Member,~IEEE,} 
Jingcai Guo,~\IEEEmembership{Member,~IEEE,} 
Xiaocheng Lu,~\IEEEmembership{Student Member,~IEEE,} \\
Song Guo,~\IEEEmembership{Fellow,~IEEE,} 
Peiran Dong,~and Jiewei Zhang
\thanks{Z. Liu, J. Guo, X. Lu, S. Guo, P. Dong and J. Zhang are with Department of Computing, The Hong Kong Polytechnic University, Hong Kong SAR., China (e-mail: ziming.liu@connect.polyu.hk; jc-jingcai.guo@polyu.edu.hk; xiaoclu@polyu.edu.hk; song.guo@polyu.edu.hk; peiran.dong@connect.polyu.hk; jiewei.zhang@connect.polyu.hk).}
}

\markboth{Under Review}%
{Shell \MakeLowercase{\textit{et al.}}: A Sample Article Using IEEEtran.cls for IEEE Journals}


\maketitle

\begin{abstract}
This paper investigates a challenging problem of zero-shot learning in the multi-label scenario (MLZSL), wherein, the model is trained to recognize multiple unseen classes within a sample (e.g., an image) based on seen classes and auxiliary knowledge, e.g., semantic information. 
Existing methods usually resort to analyzing the relationship of various seen classes residing in a sample from the dimension of spatial or semantic characteristics, and transfer the learned model to unseen ones. 
But they ignore the effective integration of local and global features. That is, in the process of inferring unseen classes, global features represent the principal direction of the image in the feature space, while local features should maintain uniqueness within a certain range. This integrated neglect will make the model lose its grasp of the main components of the image. Relying only on the local existence of seen classes during the inference stage introduces unavoidable bias.
In this paper, we propose a novel and effective group bi-enhancement framework for MLZSL, dubbed \textit{GBE-MLZSL}, to fully make use of such properties and enable a more accurate and robust visual-semantic projection. Specifically, we split the feature maps into several feature groups, of which each feature group can be trained independently with the Local Information Distinguishing Module (LID) to ensure uniqueness. Meanwhile, a Global Enhancement Module (GEM) is designed to preserve the principal direction. Besides, a static graph structure is designed to construct the correlation of local features.
Experiments on large-scale MLZSL benchmark datasets \textit{NUS-WIDE} and \textit{Open-Images}-v4 demonstrate that the proposed \textit{GBE-MLZSL} outperforms other state-of-the-art methods with large margins.
\end{abstract}


\section{Introduction}
\IEEEPARstart{W}{ith} the continuous advances in science and technology, the ability and ways of human beings to obtain information have been greatly improved. Also, with the emergence of smart devices, the image information has been greatly expanded, and the magnitude of the database is also expanding. 
In the era of big data, human beings are faced with the problem of how to deal with large-scale data for the first time. 
In recent years, deep learning classification networks~\cite{simonyan2014very, szegedy2015going} driven by computational intelligence have sprung up and made rapid progress. 
Image classification datasets designed for single labels in the past are less challenging for deep learning models. Since an arbitrary image may contain multiple objects, so the task of multi-label image classification~\cite{gong2013deep, yu2014large, weston2011wsabie} has come into view because the task is challenging, and it is closer to the real life of human. It has the following challenges compared to traditional single-label image classification tasks. First of all, the existence of multiple labels needs to consider the relevance of different labels in the same image. Secondly, the expansion of the number of classes increases the difficulty of classification. 
\begin{figure}[t]
\centering
\includegraphics[width=0.95\linewidth]{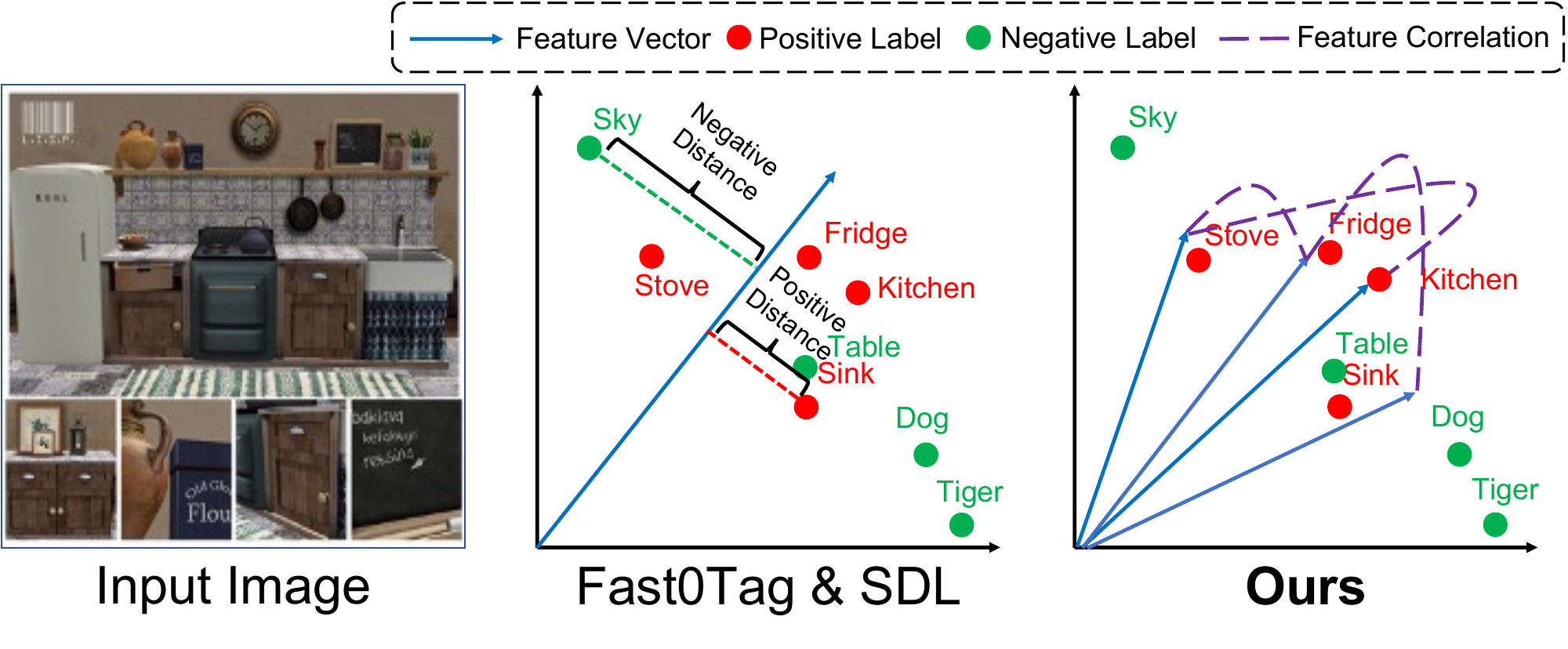}
\caption{A comparison between our GBE-MLZSL and past methods. We can conclude that for input images, compared to Fast0Tag and SDL, which only rely on ranking the distance between different classes and the principal direction of feature vectors, our method is more in line with the exploration of multi-label practical problems by generating multiple sub-vectors and constructing relationships among them.}
\label{fig:framework}
\end{figure}

With the continuous improvement of deep neural networks, multi-label image classification models based on deep learning have emerged~\cite{wang2016cnn, durand2019learning, gao2021learning, feng2019collaboration}. Their research focuses on how to effectively process the context information in the image, so as to better mine the rich information contained in the image, then match each label of the image with the semantic information, and finally transform the multi-label problem into multiple binary classification problem. Or build a graph model or other topological models to handle complex dependencies between labels~\cite{chen2019multi}. However, the former ignores the possible topological relationships between labels or object proposals, while the latter cannot obtain high-quality semantic information only with graph networks. In recent years, researchers have realized that the local features of images can be associated with labels, and at the same time, the attention mechanism has been introduced to allow the model to focus more selectively on information-rich regions and has achieved remarkable results~\cite{cheng2021mltr}. Some works enhance the model's perception of local features by introducing a spatial attention mechanism. 

However, the above multi-label image classification methods all have a common problem, that is, a large amount of data and calculations are required to achieve better results. At the same time, these models cannot perceive categories that were not present during training. Therefore, zero-shot learning (ZSL) has become a solution to this problem~\cite{akata2016multi, deutsch2017zero, frome2013devise}. 
ZSL is essentially a multi-modal transfer learning, that is, an algorithm that predicts unseen classes during testing through the learning of seen classes. Therefore, it is different from the tasks and challenges faced by traditional supervised learning. In fact, existing ZSL models have achieved substantial success in single object prediction~\cite{kodirov2017semantic, xian2017zero, guo2020novel}. This is due to the fact that the detection of seen classes by the deep neural network has become more accurate. At the same time, the more common intermediate information for knowledge transfer is mainly visual features or semantic word vectors. 

Compared with traditional ZSL, the task challenge of generalized ZSL is more difficult, and it is closer to the actual application scenario. Generalized ZSL means that not only unseen classes need to be predicted, but also all seen classes should be detected during testing. There is also a difficulty in generalized ZSL, that is, the classifier tends to seen classes during the test process, thereby weakening the predictive performance of unseen classes. 
The above studies on ZSL and generalized ZSL focus more on different attributes of the same object. But in practical problems, there are often multiple seen objects and unseen objects in a picture. At this time, the traditional ZSL model will cause serious performance loss, and even cannot work effectively. Therefore, multi-label zero-shot learning (MLZSL) has entered the research field as a more complex and practical research direction. 

There have been some works focusing on MLZSL tasks and obtained some achievements with decisive progress in recent years~\cite{norouzi2014zero, mensink2014costa, zhang2016fast}. 
Among them, Zhang~\textit{et al.}~\cite{zhang2016fast} calculates the principal direction of its feature vectors for each test image, and then sorts the similarity of labels that exist near that main direction. On the basis of it, Ben~\textit{et al.}~\cite{ben2021semantic} modifies the loss function, so that the model has the ability to identify hard-negative labels. At the same time, it also improves the diversity of semantic information extraction. However, this method only vaguely extracts the global information of the image, resulting in inaccurate principal directions of its feature vectors. 
\textit{LESA}~\cite{huynh2020shared} and \textit{BiAM}~\cite{narayan2021discriminative}, on the other hand, performs multi-label prediction tasks from another perspective, which strengthens the feature response of each class through attention mechanisms. \textit{LESA}~\cite{huynh2020shared} divides the input image into patches for attention extraction, then shares the obtained attention information into a unified attention region, and considers it as the main response area in the image. This approach provides a weak connection between different labels, but it does not highlight the characteristics of a single class. \textit{BiAM}~\cite{narayan2021discriminative} provides another perspective, which highlights the response of a single class. However, the model did not establish the relationship between classes and made local and global features independent of each other.

In response to the problems encountered in the MLZSL method mentioned above, the first problem is that previous methods did not effectively utilize the relationship between local and global features of images. For example, \textit{BiAM}~\cite{narayan2021discriminative} simply concatenates global and local features, and then performs the classification part. In the multi-label environment, global features are the most important for determining the main components in an image. However, \textit{SDL}~\cite{ben2021semantic} and \textit{Fast0Tag}~\cite{zhang2016fast} only focus on the principal vector of the image, ignoring the supplement of local features to global information. As shown in Figure~\ref{fig:framework}, `table' is actually closer to the principal vector than `sink', and `table' is not the correct seen label. Misjudgments like this often occur until our model chooses to rely on the principal vector of the image feature and uses local information to assist, which can help reduce misjudgments. At the same time, the generated multiple vectors contain both local information and global features, allowing the model to classify labels around the principal vector in the classifier. In addition, we construct an association graph for each vector and directly establish the relationship between labels.

The main contributions of this paper can be summarized as the following three-fold:
\begin{enumerate}
    \item We propose a new multi-layer feature enhancement fusion module (ML-FEF). This module effectively integrates feature responses and improves the information abundance of subsequent prediction modules.
    \item We design a global enhancement module to ensure sufficient reinforcement of global features, and a Local Information Distinguishing (LID) module to thoroughly explore the local features. Both modules work together on the fused features.
    %
    \item We introduce a global-local association graph to integrate local and global features in which connections can be constructed for features of different groups.
\end{enumerate}

\begin{figure*}
    \centering
    \includegraphics[width=0.95\textwidth]{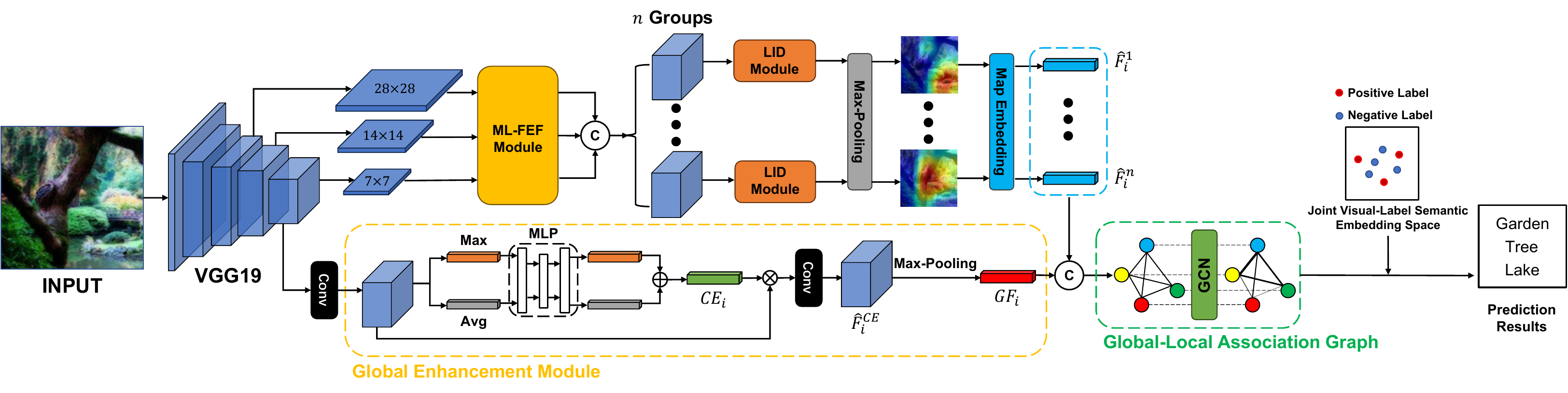}
    \caption{\textbf{Pipeline for the proposed GBE-MLZSL} (zoom in for a better view).}
    \label{fig2}
    \vspace{-5px}
\end{figure*}

\section{Related Work}
\subsection{Multi-Label Classification}
Graph neural networks (GNNs) have appeared as a promising technique for exploring label relationships, resulting in significant success in multi-label classification tasks. Specifically, \cite{chen2019multi} employs directed graphs to represent object labels, and incorporates graph convolutional networks (GCN) to map node embeddings into classifiers. In addition, Convolution-based multi-label classifiers, such as those proposed by~\cite{wang2016cnn, gong2013deep, yu2014large, chen2019multi, weston2011wsabie, durand2019learning, feng2019collaboration, gao2021learning}, can learn label characteristics from image spatial information, leading to the design of new classifiers. \cite{gao2021learning} further proposes a two-pipeline framework that exploits both local and global information, and introduces a multi-class regional attention module to bridge between these streams. An important challenge in multi-label classification is the high cost of annotating samples for each category. To solve this problem, Cole~\textit{et al.}~\cite{cole2021multi} and subsequent work\cite{kim2022large, ke2022hyperspherical,verelst2023spatial} have introduced single positive multi-label (SPML) classification, aiming at developing efficient multi-label image classifiers with merely one positive label for each image. However, the disproportion between abundant negative samples and scarce positive samples may adversely affect model performance. To overcome the label bias problem, Huang~\textit{et al.}~\cite{huang2023asymmetric} present an asymmetric polynomial loss function for fine-tuning the polynomial coefficients and asymmetric focusing parameters for various tasks and models. The designed loss function can alleviate the imbalance between positive and negative instances through an asymmetric focusing mechanism. This mechanism establishes a strong connection between the polynomial coefficients and the asymmetric focusing parameters for both positive and negative classes. It is worth noting that these methods only recognize only seen classes and cannot generalize to unseen classes.

\subsection{Zero-Shot Learning}
Zero-shot learning addresses the issue of existing models not being able to fit into unseen classes. Among the areas of zero-shot learning, single-label zero-shot learning ~\cite{farhadi2009describing,frome2013devise,shigeto2015ridge,bucher2016improving,zhang2016zero,changpinyo2016synthesized,kodirov2017semantic,zhu2018generative,li2020learning,zhang2019tgg,long2018pseudo,guo2020novel,liu2022towards,lu2023decomposed,guo2023graph} is the most widely studied and relatively straightforward. In practical applications, the focus of zero-shot learning models is primarily on identifying the primary semantic information of training images and exploiting the semantic relationship between seen and unseen categories for accurate predictions, whereby word~\cite{xian2017zero, frome2013devise, socher2013zero, xian2016latent} and attribute~\cite{lampert2014attribute, akata2016label, lampert2009learning, 7883945, ji2019attribute} vectors are used to represent such relationships. The semantic information generated can be deduced from seen to unseen labels by assessing the similarity between their respective relationship vectors. Chen~\textit{et al.}~\cite{chen2021mitigating} introduced a generative flow framework and a combinatorial strategy to tackle some of the common issues with zero-shot learning, such as semantic inconsistency, variance collapse, and structural disorder. Gune~\textit{et al.}~\cite{gune2020generalized} developed a method that utilizes generated visual proxy samples to simulate the average entropy of the label distribution for the unseen class. However, most models in zero-shot learning make predictions by learning a single representation of the image, which is not effective for practical problems in multi-label environments. This limitation makes single-label zero-shot learning difficult to generalize to multi-label scenarios, especially when multiple unseen labels need to be predicted. Chen~\textit{et al.}~\cite{9768177} proposed a graph-navigated dual attention network to jointly learn local and explicit global embeddings with a region-guided attention network and region-guided graph attention network. A self-calibration mechanism is designed to improve the visual-semantic interaction and prevent the overfitting of unseen classes. Feng~\textit{et al.}~\cite{feng2020transfer} presents a resource-efficient transfer-increment mechanism for generalized zero-shot learning. It utilizes a linear generative model with dual knowledge sources to synthesize exemplars for unseen classes and introduces two training modes, IWM and IOM, for incremental learning. IOM yields the highest harmonic mean results, while IWM excels in recognizing seen classes.

\subsection{Multi-Label Zero-Shot Learning}

Multi-label zero-shot learning presents a more complex challenge than single-label learning due to the unpredictability of the number of labels assigned to each image, thus demanding the model's capacity to weigh multiple unseen labels simultaneously. 
Several methods have emerged to address the challenge. Norouzi \textit{et al.} \cite{norouzi2014zero} developed a model for multi-label zero-shot learning by explicitly partitioning the image and semantic embedding spaces and using a convex combination of label embedding vectors to map the image. Zhang \textit{et al.} \cite{zhang2016fast} proposed a more general and fast model based on word vectors of ranked relevant labels. Lee \textit{et al.} \cite{lee2018multi} posited that knowledge graphs offer a framework for linking different labels in multi-label environments. Attention-based methods such as LESA \cite{huynh2020shared} and Narayan \textit{et al.} \cite{narayan2021discriminative} have shown promising results. Their approaches leverage the attention-sharing mechanism and bi-layer attention module to focus on key areas of different labels and global context information, respectively. Although Ben \textit{et al.} \cite{ben2021semantic} utilized the diversity of semantics and embedding matrices to enhance the ability of multi-label zero-shot learning, such methods require analysis in real-world scenarios. Nevertheless, these approaches fall short of exploring the feature channels' response to various classes and only stay at the two-dimensional ($H \times W$) level. 

\section{Methods}
\subsection{Preliminary Problem Setting}
Firstly, it is necessary to clearly define the task purpose of MLZSL. Let the batch size be equal to $n$, for the images input into the model $\left \{ \left ( I_1, Y_1\right ),\dots, \left ( I_i, Y_i\right ), \dots, \left ( I_n, Y_n\right )\right \} $, where $I_i$ is the $i$-th image in the input training set, and the corresponding $Y_i$ represents the label of the input image $i$. These labels in the training set are also called `seen labels'. Like ZSL, MLZSL does not overlap the labels of the training and testing sets in terms of label distribution. Let's define the set of all labels in the dataset as $\mathcal{C}$, the set of seen labels as $\mathcal{C}_s$, and the set of unseen labels as $\mathcal{C}_u$. The distribution relationship of labels in the dataset can be described as $\mathcal{C} = \mathcal{C}_s\cup\mathcal{C}_u$. $\mathcal{C}_s$ is mainly used for training sets, while $\mathcal{C}_u$ mainly appears during the testing process. During the MLZSL testing process, for the input image $I_u$, the output prediction result $y_u$ should be $y_u \subset \mathcal{C}_u$. In the generalized MLZSL task, the output prediction result $y_u$ must include `seen labels', which is $y_u \subset \mathcal{C}$.
\begin{figure}[htbp]
    \centering
    \includegraphics[width=.27\textwidth]{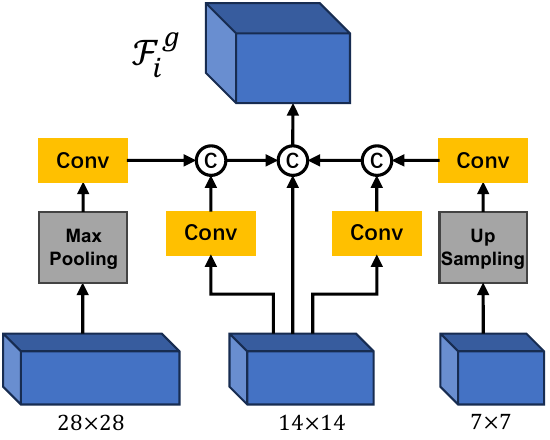}
    \caption{The structure of \textbf{Multi-Layer Feature Enhancement Fusion (ML-FEF)}.}
    \vspace{-5px}
    \label{fig:quali}
\end{figure}
\subsection{Multi-Layer Feature Enhancement Fusion}
Previous research methods, whether focused on attention based or calculating the principal direction of image feature vectors, typically only chose to use feature maps with scale $14\times 14$. This single feature layer directly leads to limitations in subsequent feature extraction and correlation. So in our proposed method, we decided to use multi-layer feature maps to process as rich features as possible, which will help us build graph relationships between labels in the future. 

Firstly, we choose VGG19 to get multi-layer feature maps. For the input image $i$, multi-layer features are defined as $\mathcal{F}_i=\{F_i^{28}, F_i^{14}, F_i^{7}\}$. The superscript indicates the size of the feature map. multi-layer aggregation is then performed on $F_i^{28}$ and $F_i^{14}$, and $F_i^{14}$ and $F_i^{7}$. 
As shown in Figure 3, the feature maps are first filtered through the convolutional layer, then down-sampled (or up-sampled) to the same size as $F_i^{14}$, and finally aggregated. The fused feature map is defined as $\mathcal{F}_i^g$. 
\begin{align}
    \mathcal{F}_i^{14,28} = F_i^{14} \odot F_i^{28},
\end{align}
\begin{align}
    \mathcal{F}_i^{14,7} = F_i^{14} \odot F_i^{7},
\end{align}
\begin{align}
    \mathcal{F}_i^g = F_i^{14} \oplus {F}_i^{14,7} \oplus \mathcal{F}_i^{14,28}.
\end{align}
Among them, $\odot$ represents the multi-layer feature fusion operation, and $\oplus$ represents the concatenate between different feature maps. 
\begin{figure}[htbp]
    \centering
    \includegraphics[width=.47\textwidth]{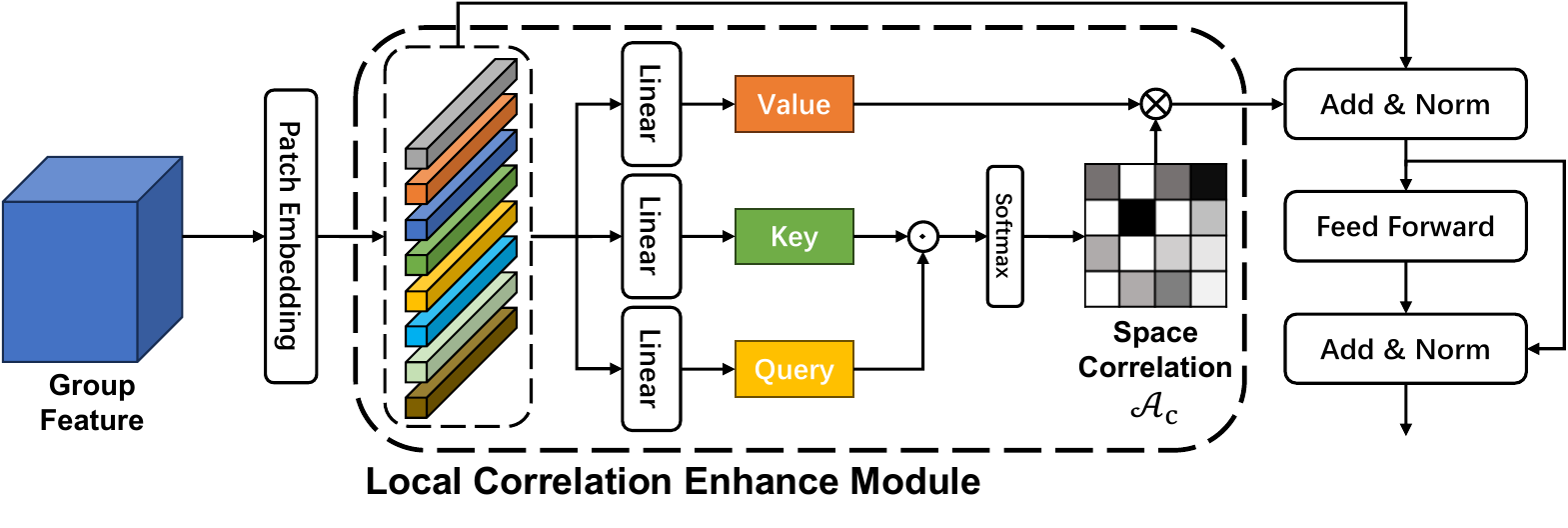}
    \caption{The structure of \textbf{Local Information Distinguishing Module}.}
    \vspace{-5px}
    \label{fig:quali}
\end{figure}

\subsection{Local Information Distinguishing Module}
Since we need to predict multiple unseen labels at the same time, the information represented by different channels of the feature map is required to be more differentiated. Referring to the currently popular strategy of grouping, we also group the generated feature maps $\mathcal{F}_i^g$ into $\mathcal{F}_i^g=\{F_i^{g,1},\dots, F_i^{g,n}\}$, where $n$ represents the number of groups. The number of channels of each sub-feature map is $d_w$, where the size of $d_w$ is equal to the length of word embedding.

For the features after grouping, we need to fully extract their local information. In this section, in order to capture the long-range dependencies between different regions, we choose self-attention instead of convolution operations. The specific reason is that the target we are dealing with is a multi-label environment, so the label relationship needs to be considered. The feature space processed by the convolution itself has its limitation in proceeding long-range relationships, so we choose self-attention instead of convolution. 
Taking the $m$-th sub-feature map $F_i^{g,m}$ as an example, we first divide it into patches through an embedding layer and get $F_i^{a,m}$. $F_i^{a,m}$ is then input into the local correlation enhance module.
\begin{equation}
    \mathbf{Q}=W^QF_i^{a,m}, \quad \mathbf{K}=W^KF_i^{a,m}, \quad \mathbf{V}=W^VF_i^{a,m},
\end{equation}
where $W^{(\cdot)}$ means the fully convolution layer. Next, to obtain the spatial correlation matrix $\mathcal{A}_c$, then we let $\mathbf{Q}$ and $\mathbf{K}$ perform a dot product operation and go through a softmax to obtain the spatial correlation matrix $\mathcal{A}_c\in \mathbb{R}^{HW\times HW}$. After that, we can get the enhanced local feature attention map by do the dot product between $\mathcal{A}_c$ and $\mathbf{V}$. Finally, we add the enhanced attention map with the input $F_i^{a,m}$ to build residuals and get the final output $\widehat{F}_i^{a,m}\in \mathbb{R}^{HW\times d_w}$:
\begin{equation}
    \mbox{Attention}(\mathbf{Q}, \mathbf{K}, \mathbf{V}) = \mbox{softmax}(\underbrace{ \mathbf{K} \cdot \mathbf{Q}}_{\mathcal{A}_c}) \cdot \mathbf{V},
\end{equation}

\begin{equation}
    \widehat{F}_i^{a,m} = F_i^{a,m} + \mbox{Attention}(\mathbf{Q}, \mathbf{K}, \mathbf{V}).
\end{equation}
Then $\widehat{F}_i^{a,m}$ passes through a Feed-Forward layer to perform nonlinear transformation on the features, so that the model can output more expressive results. The significance of the local information distinguishing module is to allow the model to generate different feature vectors, highlighting the diversity, and avoiding the neglect of detailed information by using the single principal vector. Next, we use the max-pooling operation to highlight the main semantic representation of the $\widehat{F}_i^{a,m}$ and generate enhanced local semantic information $\widehat{F}_i^m$.

\subsection{Global Enhancement Module} 
Usually, as the feature extraction network deepens, the obtained features are more advanced and abstract. Past models~\cite{zhang2016fast,ben2021semantic} only use the deepest features to represent the global features of an image. Therefore, in this section, we still retain the way of extracting global features. On the contrary, compared with the past methods what directly use global features, we need to enhance and select them. 
Firstly, we perform channel enhancement operations for the deep features. 
In the channel enhancement module, we first extract the channel information of the deep features through max-pooling and average-pooling in the spatial axis. The steps are shown in the following function:
\begin{align}
    F_i^{avg-spa} = Avg-Spatial\left(F_i^{7}\right),
\end{align}
\begin{align}
    F_i^{max-spa} = Max-Spatial\left(F_i^{7}\right).
\end{align}
Then generating max-pooling feature and average-pooling feature as the input of multi-layer perceptron (MLP) with one hidden layer. The channel enhancement is computed as:
\begin{align}
    CE_i = MLP\left(F_i^{avg-spa}\right) + MLP\left(F_i^{max-spa}\right).
\end{align}
Then we perform channel-wise multiplication with the group feature to generate the new feature $F_{CA}^{i, 1}$.
\begin{equation}
    F^{CE}_i = CE_i \otimes F_i^7 = \left[CE_i^1\left(F^7_{i, 1}\right),\dots,CE_i^n\left(F^7_{i, n}\right)\right],
\end{equation}
where $F^7_{i, n}$ refers to the $n$-th channel of the input feature map $F_i^7$. Next, for the channel-enhanced global features, we perform convolutional screening to retain the channel that best represents the global information to ensure the compactness of the global information $\widehat{F}^{CE}_i$. Finally, we gain our global semantic information $GF_i$ through a max-pooling operation.

\subsection{Global-Local Association Graph} 
In the previous two sections, we obtained enhanced local features and global features respectively. In order to enhance the accuracy of prediction and improve the correlation between semantic information generated by features, we proposed the Global-Local Association Graph. Unlike multi-label classification models, zero-shot learning is a prediction task. Therefore, dynamic graphs cannot be designed like classification models, because the relationship between each local feature is important. We chose to build a fully connected static graph to describe this relationship. 

First, we combine each enhanced local semantic information with global semantic information. The purpose of this approach is to ensure that our model achieves the diversity of semantic information without deviating from the principal vector. The single-layer static GCN is simply defined as: 
\begin{equation}
    \mathbf{V}_i = \left[(\widehat{F}_i^1;GF_i), (\widehat{F}_i^2;GF_i)\dots, (\widehat{F}_i^n;GF_i)\right],
\end{equation}

\begin{equation}
    \mathcal{S}_i = LReLU(\mathbf{A}_s\mathbf{V}_i\mathbf{W}_s),
\end{equation}
where the activation function $LReLU(\cdot)$ is LeakyReLU, the $\mathbf{A}_s$ represents the affinity matrix, $\mathbf{W}_s$ represents the state-update parameters of each node in the static graph. Finally, we can get the final semantic vector group $\mathcal{S}_i\in \mathbb{R}^{n\times d_w}$. It has the consistency of global information and the diversity of local features at the same time, and is suitable as an input for the prediction stage of unseen classes.

\subsection{Loss Function}
Our loss function is inspired by ~\cite{zhang2016fast,ben2021semantic}. In the training stage, for each input image $i$, the semantic vector group of the corresponding image output by the model is $\mathcal{S}_i \in \mathbb{R}^{n\times d_w}$. This semantic vector group is used to predict unseen classes during the testing process. Therefore, the basic calculation process of the loss function includes the judgment of class similarity. It is necessary to try to improve the ranking of the classes that appear in the input image and reduce the similarity ranking of the classes that do not appear, 
\begin{equation}
    \tau_{jk}=\max\left(n_j\mathcal{S}_i\right) - \max\left(p_k\mathcal{S}_i\right),
\end{equation}
where $n_j$ is the word vector of the seen classes that do not appear, and $p_k$ is the word vector of the seen classes that appear in the input image. In order to maximize the distance between classes that appear and those that do not, we use the maximum value for calculation. At the same time, this approach can ensure sufficient semantic diversity. Therefore, one of the main components of the loss function, namely the ranknet loss~\cite{zhang2016fast}, is shown in the following function:
\begin{equation}
    \mathcal{L}_{rank} = \alpha\sum_j\sum_k\log \left(1+e^{\tau_{jk}}\right),
\end{equation}
where $\alpha=\left(\left|T\right|\left|\bar{T}\right|\right)^{-1}$. $\left|T\right|$ and $\left|\bar{T}\right|$ denote the number of seen classes that appear or not, respectively. The hyper-parameter $\alpha$ is used to normalize the ranknet loss. 

In addition, in order to reduce the difficulty of the model in predicting hard-positive samples with a high diversity of labels, we suggest introducing a new weight like~\cite{ben2021semantic}. Due to the uniqueness of multi-label tasks, each image contains a different number of labels. When a large number of classes appear in the image, it is necessary to make the model sensitive to label diversity and have a tendency towards hard-positive tasks during the training process.
\begin{equation}
    \omega = 1 + \sum_i var(Y_i). 
\end{equation}
In addition, the loss function of the model introduces the second main component, namely the regularized loss function. Its purpose is to construct correlations between input semantic vectors.
\begin{equation}
    \mathcal{L}_{reg} = \left \| \sum_n var(\mathcal{S}_i^n) \right \|_1.
\end{equation}
Finally, our complete loss function $\mathcal{L}_{final}$ is as follows:
\begin{equation}
    \mathcal{L}_{final} = \frac{1}{N}\sum_{i=1}^N\left(w \cdot (1-\lambda)\mathcal{L}_{rank}(\mathcal{S}_i, Y_i) + \lambda \mathcal{L}_{reg}(\mathcal{S}_i)\right),
\end{equation}
where $\lambda$ is the weight of the regularized loss function, and $N$ is the batch size.

\begin{table*}[htbp]
  \centering
  \caption{Comparison of different models for multi-label ZSL and GZSL tasks on the \textbf{NUS-WIDE} dataset. The ``P(K)'' refer to the top-K ``Precision'' and ``R(K)'' refers to the top-K ``Recall''. Best results are shown in bold. }
    \begin{tabular}{lc|cccccc|c}
    \toprule
    \textbf{Method} & \textbf{Task} & \textbf{P (K = 3)} & \textbf{R (K = 3)}  & \textbf{F1 (K = 3)} & \textbf{P (K = 5)} & \textbf{R (K = 5)} & \textbf{F1 (K = 5)} & \textbf{mAP} \\
    \midrule
    \multirow{2}[2]{*}{\textit{CONSE}~\cite{norouzi2014zero}}  & ZSL   & 17.5 & 28.0   & 21.6  & 13.9 & 37.0 & 20.2 & 9.4\\
                               & GZSL  & 11.5 & 5.1     & 7.0   & 9.6 & 7.1  & 8.1  & 2.1\\
    \cmidrule{2-9}  
    \multirow{2}[2]{*}{\textit{LabelEM}~\cite{akata2015label}}& ZSL   & 15.6  & 25.0  & 19.2 & 13.4  & 35.7  & 19.5 & 7.1\\
                               & GZSL  & 15.5  & 6.8  & 9.5 & 13.4  & 9.8  & 11.3 & 2.2\\
    \cmidrule{2-9}  
    \multirow{2}[2]{*}{\textit{Fast0Tag}~\cite{zhang2016fast}}& ZSL  & 22.6  & 36.2  & 27.8 & 18.2  & 48.4  & 26.4 & 15.1 \\
                                & GZSL & 18.8  & 8.3  & 11.5 & 15.9  & 11.7  & 13.5 & 3.7 \\
    \cmidrule{2-9}  
    \multirow{2}[2]{*}{Attention per Label~\cite{kim2018bilinear}}& ZSL   & 20.9  & 33.5  & 25.8 & 16.2  & 43.2  & 23.6 & 10.4 \\
                                           & GZSL  & 17.9  & 7.9   & 10.9 & 15.6  & 11.5  & 13.2 & 3.7 \\
    \cmidrule{2-9}
    \multirow{2}[2]{*}{\textit{Deep0Tag}~\cite{rahman2019deep0tag}}& ZSL   & 25.7  & \bf{43.8}  & 32.4 & 17.2  & 48.9  & 25.5 & - \\
                                           & GZSL  & \bf{33.8}  & 13.1   & 18.9 & 23.7  & 15.3  & 18.5 & - \\
    \cmidrule{2-9}  
    \multirow{2}[2]{*}{Attention per Cluster~\cite{huynh2020shared}} & ZSL   & 20.0  & 31.9  & 24.6 & 15.7  & 41.9  & 22.9 & 12.9 \\
                                              & GZSL  & 10.4  & 4.6  & 6.4 & 9.1  & 6.7  & 7.7 & 2.6 \\
    \cmidrule{2-9}  
    \multirow{2}[2]{*}{\textit{LESA} (M = 10)~\cite{huynh2020shared}} & ZSL   & 25.7  & 41.1  & 31.6 & 19.7  & 52.5  & 28.7 & 19.4 \\
                                      & GZSL  & 23.6  & 10.4  & 14.4 & 19.8  & 14.6  & 16.8 & 5.6 \\
    \cmidrule{2-9}
    \multirow{2}[2]{*}{\textit{BiAM}~\cite{narayan2021discriminative}} & ZSL   & 26.0  & 41.6  & 32.0 & 20.2  & 53.9  & 29.4 & 25.8 \\
                                      & GZSL  & 25.2  & 11.2  & 15.5 & 21.8  & 16.0  & 18.5 & 8.9 \\
    \cmidrule{2-9} 
    \multirow{2}[2]{*}{\textit{SDL}~\cite{ben2021semantic}} & ZSL   & 23.4  & 37.1  & 28.7 & 17.9  & 48.9  & 26.2 & 24.0 \\
                                      & GZSL  & 25.9  & 11.4  & 15.8 & 21.1  & 17.0  & 18.8 & 9.5 \\
    \cmidrule{2-9} 
    \multirow{2}[2]{*}{\bf{Ours}}& ZSL   & \bf{26.9}     & 42.9     & \bf{33.1}  &  \bf{23.4}     & \bf{54.3}     & \bf{32.7} & \bf{28.7}\\
                                    & GZSL  & 31.2     & \bf{13.9}     & \bf{19.2} & \bf{25.4}     & \bf{17.6}     & \bf{20.8} & \bf{10.3} \\
    \bottomrule
    \end{tabular}%
  \label{tab1}%
\end{table*}

\section{Experiments}
\subsection{Experimental Setup}
\noindent\textbf{Datasets:} The \textit{NUS-WIDE} dataset~\cite{chua2009nus} contains approximately 270,000 images and a total of 1,006 labels. Among them, 81 labels manually annotated by humans will serve as labels for `unseen classes'. At the same time, these labels will also serve as `ground-truth' labels in the multi-label classification task. The remaining 925 labels were automatically extracted from Flickr users' manual annotations of these images, where they will be used as labels for `seen classes'. This setting is similar with~\cite{huynh2020shared, ben2021semantic}. 
\noindent Another dataset is called the \textit{Open-Images}-V4 dataset, which is much larger than the \textit{NUS-WIDE} dataset. This dataset contains approximately 9.2 million images, of which approximately 9 million are used as the training set. The training set contains a total of 7,186 labels, ensuring that each label appears at least 100 times in the training set. These will be considered as labels for `seen classes'. In addition, the dataset also contains 125,456 test images and 400 `unseen classes' labels. These labels are derived from the other 400 most frequent labels that did not appear in the training set, which appeared at least 75 times. The setting of \textit{Open-Images}-V4 dataset is similar with~\cite{huynh2020shared, narayan2021discriminative}.

\noindent\textbf{Evaluation Metrics:} In order to ensure the unbiased comparison and scientific evaluation of the metrics themselves using our proposed method and comparison method, we will use the two most commonly used evaluation metrics in MLZSL and Multi-label Classification tasks, namely the mean Average Precision (mAP) and F1-Score~\cite{veit2017learning, huynh2020shared}. 
Among them, F1-Score is the harmonic mean of the precision and recall. \textit{top-K} F1-Score is an evaluation metric used in classification tasks to measure the accuracy of the model in predicting labels. 
mAP is a class-wise evaluation indicator, which is used to reflect the accuracy of unseen label retrieval of the image. 

\noindent\textbf{Implementation Details:} As for the selection of backbone network, we choose the VGG19~\cite{simonyan2014very} network that is pre-trained on the \textit{ImageNet} dataset ~\cite{deng2009imagenet} as our backbone network. At the same time, our method will use feature maps with scales of $28\times 28$, $14\times 14$, and $7\times 7$ for comparison. In the ablation experiment, we will use feature maps with the same scale as the comparison method ~\cite{huynh2020shared, narayan2021discriminative} as a reference.
Unlike the comparison method that uses multi-stage training, our method is completely end-to-end training, saving a lot of training time and making the model more concise.

\noindent We choose the Adam optimizer~\cite{kingma2014adam}as the model's optimizer, which is suitable for large-scale datasets and requires less memory. The weight decay of the Adam optimizer is set to $4e^{-3}$. For the experiments of all the models in the \textit{NUS-WIDE} dataset, the entire training process requires a total of 20 epochs with a batch size of 96, and the initial learning rate is set to $1e^{-4}$, and then decreases by $\frac{1}{10}$ at the 7-th and 14-th epoch, respectively. 
In the experiments of the \textit{Open-Images}-V4 dataset, the number of epochs in the training process is set to 7. This is consistent with the settings for other comparison methods. our optimizer's decay rate, model's learning rate, batch size are remain the same.

\noindent\textbf{Baselines:} 
In terms of baseline methods, we need to compare our proposed new model with all state-of-the-art MLZSL models based on deep neural networks in recent years. The baseline method covers all research directions, including but not limited to Generative Adversarial Network (GAN), attention mechanism, principal direction of feature vectors, etc.
These comparison methods include:
\textit{CONSE}~\cite{norouzi2014zero}, 
\textit{LabelEM}~\cite{akata2015label}, 
\textit{Fast0Tag}~\cite{zhang2016fast}, 
\textit{Attention per Label}~\cite{kim2018bilinear}, 
\textit{Deep0Tag}~\cite{rahman2019deep0tag},
\textit{LESA Attention per Cluster}~\cite{huynh2020shared}, 
\textit{LESA}~\cite{huynh2020shared}, 
\textit{BiAM}~\cite{narayan2021discriminative}.
and \textit{SDL}~\cite{ben2021semantic}. 
\noindent All comparison methods use pre-trained VGG19~\cite{simonyan2014very} as the backbone network. 

\subsection{Comparison of the MLZSL performance on NUS-WIDE}
We show the performance of both ZSL and GZSL tasks in Table~\ref{tab1}. 
Firstly, for some earlier proposed models, such as \textit{CONSE}~\cite{norouzi2014zero} and \textit{LabelEM}~\cite{akata2015label}, Their prediction mechanism is relatively backward, and there is no strong correlation between the features and semantic information extracted by the deep model, so their performance is poor. \textit{LESA}~\cite{huynh2020shared} and \textit{BiAM}~\cite{narayan2021discriminative}, as two models with the help of spatial attention mechanism, they are inspired by the multi-label classification methods, and they pay attention to the attention sharing between labels and the personalized attention mechanism respectively, have made enough progress. \textit{Deep0Tag}~\cite{rahman2019deep0tag} is inspired by target detection, and improves the model's sensitivity to multi-labels by using regions of interest. This approach has achieved relatively good F1-Score results in the face of a small number of unseen labels, but as the number of predictions increases, the performance of the model suffers a great attenuation. \textit{Fast0Tag}~\cite{zhang2016fast} and \textit{SDL}~\cite{ben2021semantic}, two models that use the principal vector of the image as the prediction information, \textit{SDL}~\cite{ben2021semantic} has a certain hard split on the principal vector compared with A, which objectively improves the multi-label performance. However, our model takes into account the principal vector information and local information, so even in the case of $K=5$, the performance does not attenuate, reflecting the superior multi-label ability. At the same time, the excellent performance of GZSL also demonstrates the strong generalization ability.

\begin{table*}[htbp]
  \centering
  \caption{Comparison of different models for multi-label ZSL and GZSL tasks on the \textbf{Open-Images}-V4 dataset. The ``P(K)'' refer to the top-K ``Precision'' and ``R(K)'' refers to the top-K ``Recall''. Best results are shown in bold. }
    \begin{tabular}{lc|cccccc|c}
    \toprule
    \textbf{Method} & \textbf{Task} & \textbf{P (K = 10)} & \textbf{R (K = 10)}  & \textbf{F1 (K = 10)} & \textbf{P (K = 20)} & \textbf{R (K = 20)} & \textbf{F1 (K = 20)} & \textbf{mAP} \\
    \midrule
    \multirow{2}[2]{*}{\textit{CONSE}~\cite{norouzi2014zero}}  & ZSL   & 0.2 & 7.3   & 0.4  & 0.2 & 11.3 & 0.3 & 40.4\\
                               & GZSL  & 2.4 & 2.8     & 2.6   & 1.7 & 3.9  & 2.4  & 43.5\\
    \cmidrule{2-9}  
    \multirow{2}[2]{*}{\textit{LabelEM}~\cite{akata2015label}}& ZSL   & 0.2  & 8.7  & 0.5 & 0.2  & 15.8  & 0.4 & 40.5\\
                               & GZSL  & 4.8  & 5.6  & 5.2 & 3.7  & 8.5  & 5.1 & 45.2\\
    \cmidrule{2-9}  
    \multirow{2}[2]{*}{\textit{Fast0Tag}~\cite{zhang2016fast}}& ZSL  & 0.3  & 12.6  & 0.7 & 0.3  & 21.3  & 0.6 & 41.2 \\
                                & GZSL & 14.8  & 17.3  & 16.0 & 9.3  & 21.5  & 12.9 & 45.2 \\
    \cmidrule{2-9}  
    \multirow{2}[2]{*}{Attention per Cluster~\cite{huynh2020shared}} & ZSL   & 0.6  & 22.9  & 1.2 & 0.4  & 32.4  & 0.9 & 40.7 \\
                                              & GZSL  & 15.7  & 18.3  & 16.9  & 9.6  & 22.4  & 13.5 & 44.9 \\
    \cmidrule{2-9}  
    \multirow{2}[2]{*}{\textit{LESA} (M = 10)~\cite{huynh2020shared}} & ZSL   & 0.7  & 25.6  & 1.4  & 0.5  & 37.4  & 1.0 & 41.7 \\
                                      & GZSL  & 16.2  & 18.9  & 17.4 & 10.2  & 23.9  & 14.3 & 45.4 \\
    \cmidrule{2-9}  
    \multirow{2}[2]{*}{\textit{SDL}~\cite{ben2021semantic}} & ZSL   & 4.3  & 29.8  & 7.5 & \textbf{3.8}  & 32.8  & \textbf{6.8} & 55.0 \\
                                      & GZSL  & 27.1  & 33.6  & 30.0 & 17.4  & 38.8  & 24.0 & 66.7 \\
    \cmidrule{2-9} 
    \multirow{2}[2]{*}{\textit{BiAM}~\cite{narayan2021discriminative}} & ZSL   & 2.3  & 17.8  & 4.1 & 2.0  & 31.0  & 3.7 & 62.8 \\
                                      & GZSL  & 16.4  & 18.9  & 17.6 & 10.8  & 25.0  & 15.1 & 79.6 \\
    \cmidrule{2-9} 
    \multirow{2}[2]{*}{\bf{Ours}}& ZSL   & \bf{4.4}     & \bf{33.3}     & \bf{7.8}  & 3.6     & \bf{33.9}     & 6.5 & \bf{69.9}\\
                                    & GZSL  & \bf{32.0}     & \bf{34.4}     & \bf{33.2} & \bf{24.1}     & \bf{46.8}     & \bf{31.8} & \bf{81.4} \\
    \bottomrule
    \end{tabular}%
      \vspace{-5px}
  \label{tab2}%
\end{table*}
\subsection{Comparison of the MLZSL performance on Open-Images-V4}
Open-Images V4, as a newly emerging dataset in recent years, has a very large scale and is also used by many new methods to detect the robustness of the model when facing a large number of labels. However, this dataset also has certain shortcomings, so it cannot be used as a benchmark dataset for other experiments. Firstly, the high training cost, and secondly, it is difficult for the model to converge within the specified epochs. Table~\ref{tab2} shows the performance of ours and other competitive methods on the \textit{NUS-WIDE} test-set. For fairness, we keep the same training epoch as the comparison method. Under the ZSL task, almost all models are poor in precision, mainly because of the increase in our output predictions. But our model still achieves relatively the best results, both in terms of precision and recall. The GZSL task is very difficult, and the number of labels has been expanded to nearly 5,000. We are still 3.2\% and 7.8\% ahead of \textit{SDL}~\cite{ben2021semantic} in F1-Score, respectively.
\begin{figure}[htbp]
    \centering
    \subfigure[VGG19] {\includegraphics[width=.24\textwidth]{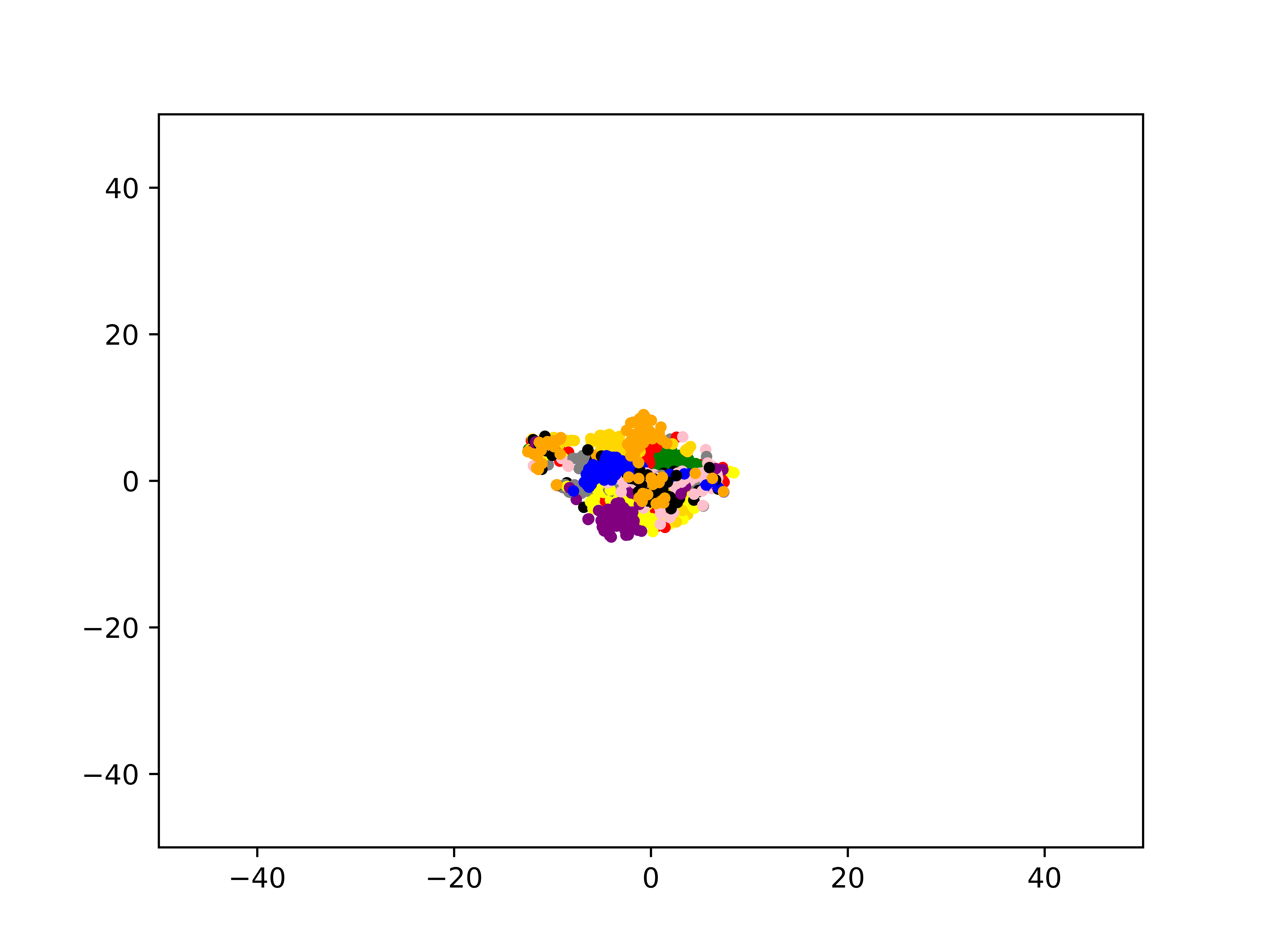}}
    \subfigure[GBE-MLZSL] {\includegraphics[width=.24\textwidth]{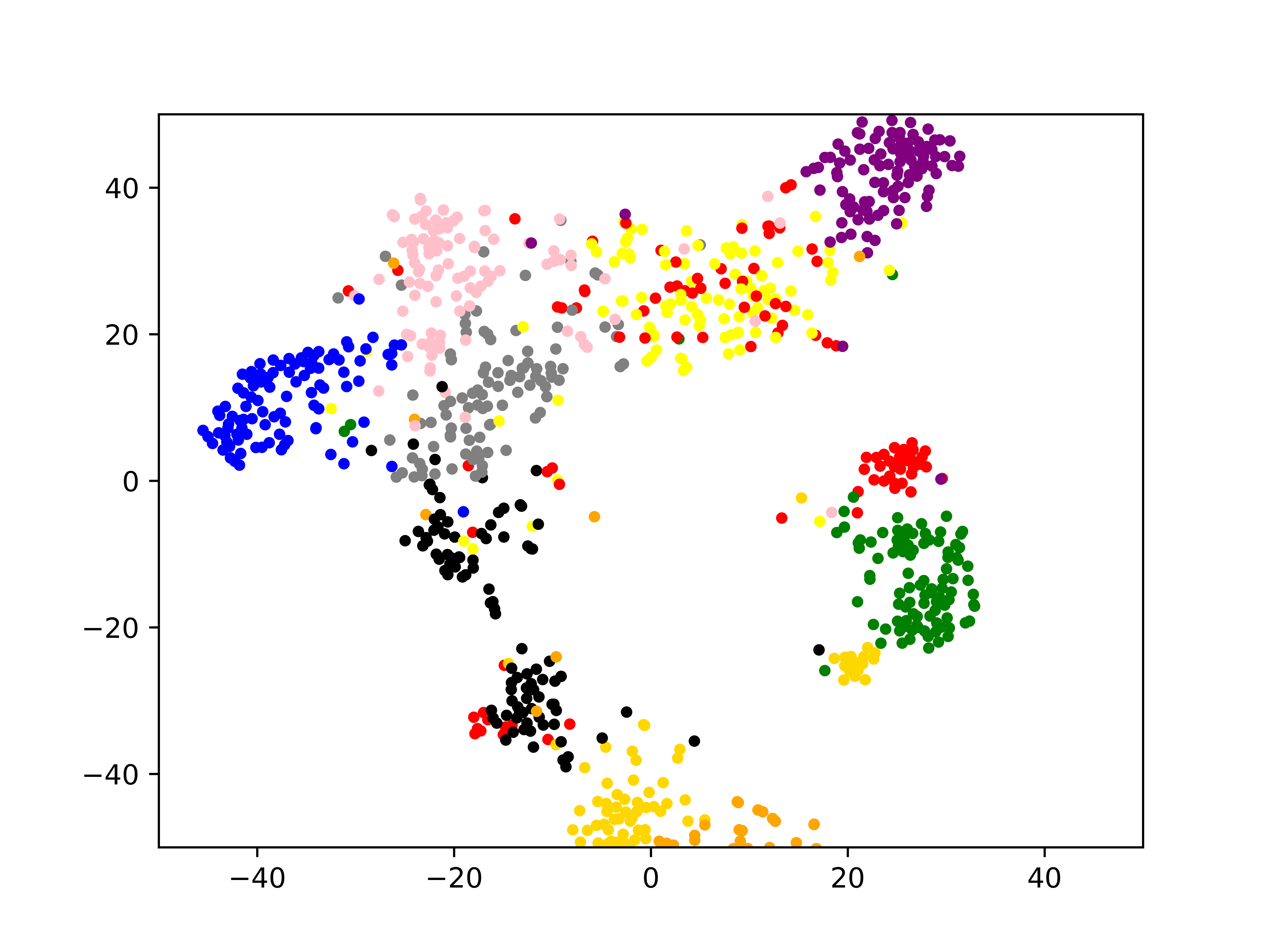}}
    \caption{\textbf{Performance of t-SNE}. It shows that after adding GBE-MLZSL, there is a very clear boundary between different classes. Zoom in for a better view.}
    \label{fig:t-SNE}
    \vspace{-10px}
\end{figure}

\begin{figure}[htbp]
    \centering
    \subfigure[$n$] {\includegraphics[width=.24\textwidth]{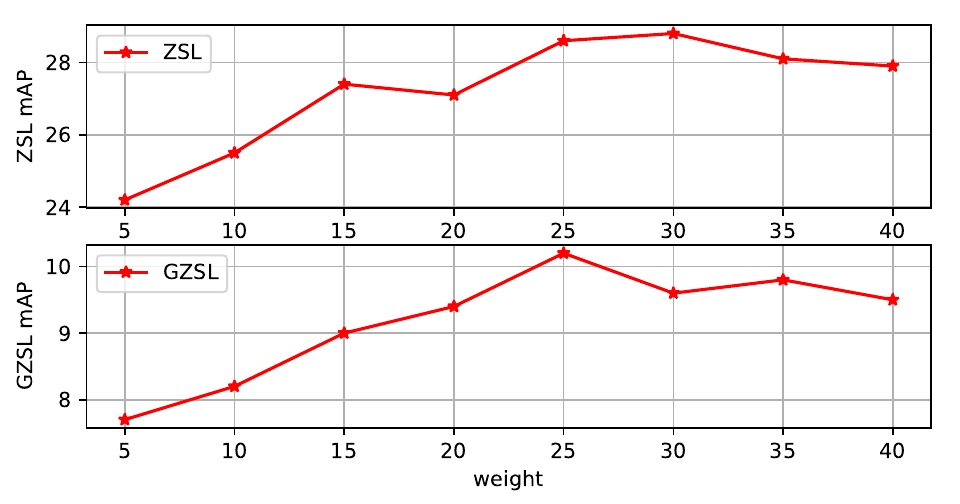}}
    \subfigure[Weights] {\includegraphics[width=.24\textwidth]{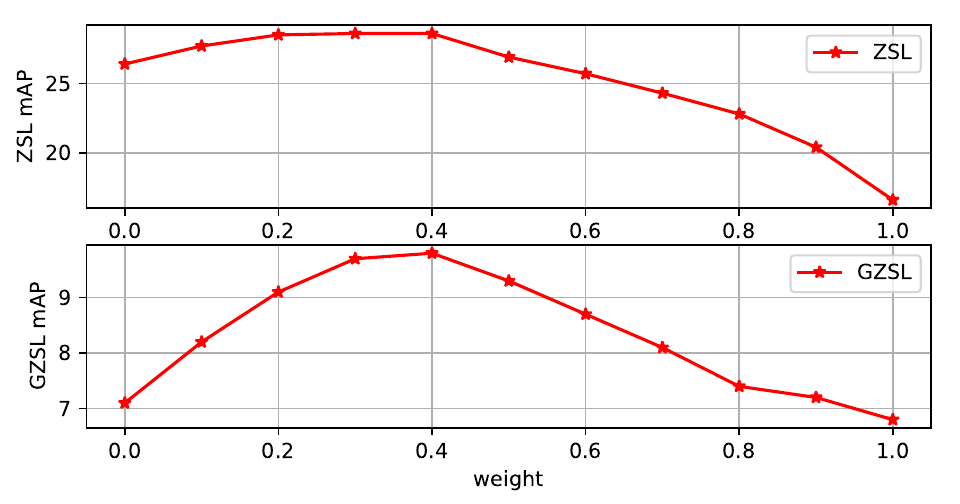}}
    \caption{\textbf{Hyper-parameter selection.} All the experiments are performed on the \textit{NUS-WIDE} test-set. The higher the mAP, the better the hyper-parameter.}
    \label{fig:Compare}
    \vspace{-5px}
\end{figure}

\begin{table}[htbp]
  \centering
  \caption{The results of \textbf{Ablation study}. The table below shows the contribution of the different modules in GBE-MLZSL. All the models are performed on the test-set of the \textit{NUS-WIDE}.}
    \begin{tabular}{cr|cccccc|c}
    \multicolumn{2}{c|}{} & a     & b     & c     &d    &e     &f     & ours \\
    \hline
    \multicolumn{2}{r|}{ML-FEF} &       & \checkmark      & \checkmark      &\checkmark      &\checkmark    &\checkmark    &\checkmark  \\
    \multicolumn{2}{r|}{LID-Module} &       &       &\checkmark       &      &\checkmark    &    &\checkmark    \\
    \multicolumn{2}{r|}{GE Module} &       &       &       &\checkmark       &    &\checkmark    &\checkmark    \\
    \multicolumn{2}{r|}{GLA Graph} &       &       &       &       &\checkmark    &\checkmark    &\checkmark    \\
    \hline
    \multirow{2}[1]{*}{\textbf{mAP}} & ZSL    & 24.1  & 25.2  & 26.9  & 26.5  & 27.7  & 27.5  &   \bf{28.7}\\
                            & GZSL    & 8.6  & 9.1  & 9.4  & 9.6  & 9.6  & 10.1  &   \bf{10.3}\\
    \end{tabular}%
  \label{tab3}%
  \vspace{-5px}
\end{table}
\begin{figure*}[htbp]
    \centering
    \includegraphics[width=.95\textwidth]{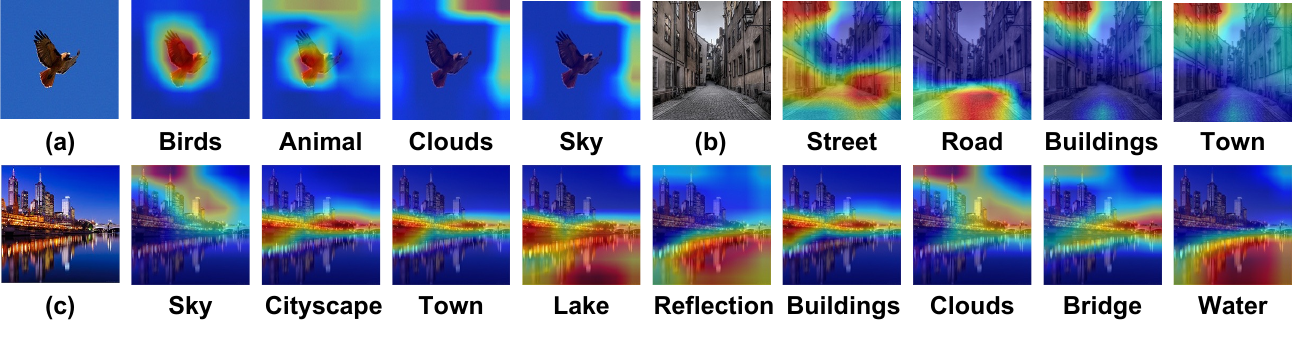}
    \caption{\textbf{Attention visualization.} where (a)(b)(c) are the attention responses of our \textbf{GBE-MLZSL} when faced with unseen labels. The red area represents the attention response of the model.}
    \vspace{-5px}
    \label{fig:attmap}
\end{figure*}

\subsection{Hyper-Parameter Selection and Ablation Study}
\noindent \textbf{Hyper-Parameter Selection:} In terms of hyper-parameter selection, our model has two hyper-parameters, $n$ and $\lambda$, as shown in the figure. First of all, we focus on the number of groups $n$. It can be seen that the model we designed does not mean that the larger the $n$, the better the performance. This is because once the features are grouped in a large number, the semantic information will be extremely scattered, which will lead to the decline of mAP results, and even adding global features will not be effective. But when the number of $n$ is too small, the semantic diversity is lost. Therefore, it can be seen from the experimental results that the model can achieve relatively best results when $n$ is 25. 

\noindent For another hyper-parameter $\lambda$, we also conducted experimental verification. It can be seen that when $\lambda=0$, the loss function of the model only includes $\mathcal{L}_{rank}$. When $\lambda=1$, the loss function of the model only contains the regular term $\mathcal{L}_{reg}$. From the perspective of the mAP results caused by the change of $\lambda$, adding the regularization term appropriately will help the convergence of the model.

\noindent \textbf{Ablation Study:} In order to verify the effectiveness of each module of our method and to test the integrity of the method, we designed an ablation study experiment. It can be seen from Table~\ref{tab3} that when the model `a' is added to ML-FEF, multi-layer features are obtained, and this change is positive for both the ZSL task and the GZSL task. In the selection of using the LID Module to enhance local features and using the GE Module to enhance the obtained global features, we can see that the addition of local features can significantly improve the ZSL task, while the enhancement of global features is more robust. This is because LID Module focuses on the mining of local features, while GE Module grasps the main information of the image and retains the generalization ability of the model. Therefore, with the subsequent addition of GLA Graph, these two modules are well integrated and a specific relationship between semantics is constructed. From the results of the ablation study, the addition of each module has fulfilled our original intention for the model.

\noindent \textbf{t-SNE Visualization:} The Figure~\ref{fig:t-SNE} shows the results of t-SNE visualization by using only VGG19 and the newly designed GBE-MLZSL. We randomly selected 10 unseen classes for display. It can be seen from the comparison that if we only use the features generated by VGG19 as semantic information to make predictions directly, the result is poor. After adding our GBE-MLZSL, the prediction boundary becomes obviously clear. This proves the correctness of our approach.

\begin{table*}[htbp]
  \centering
  \caption{State-of-the-Art models comparison on \textbf{NUS-WIDE} dataset under the task of \textbf{Multi-label Classification}. Best results are shown in bold. }
    \scalebox{1}{
    \begin{tabular}{lcccccc|c}
    \toprule
    \textbf{Method}     & \textbf{P(K=3)}  & \textbf{R(K=3)} & \textbf{F1(K=3)} & \textbf{P(K=5)}  & \textbf{R(K=5)} & \textbf{F1(K=5)} &\textbf{mAP} \\
    \midrule
    \textit{Logistic}~\cite{tsoumakas2007multi}                      & 46.1   & 57.3   & 51.1  & 34.2  & 70.8  & 46.1  & 21.6 \\
    \textit{WARP}~\cite{gong2013deep}                                & 49.1   & 61.0   & 54.4  & 36.6  & 75.9  & 49.4  & 3.1 \\
    \textit{WSABIE}~\cite{weston2011wsabie}                          & 48.5   & 60.4   & 53.8  & 36.5  & 75.6  & 49.2  & 3.1 \\
    \textit{Fast0Tag}~\cite{zhang2016fast}                           & 48.6   & 60.4   & 53.8  & 36.0  & 74.6  & 48.6  & 22.4 \\
    \textit{CNN-RNN}~\cite{wang2016cnn}                              & 49.9   & 61.7   & 55.2  & 37.7  & 78.1  & 50.8  & 28.3 \\
    One Attention per Label~\cite{kim2018bilinear}          & 51.3   & 63.7   & 56.8  & 38.0  & 78.8  & 51.3  & 32.6 \\
    One Attention per Cluster (M = 10)  & 51.1   & 63.5   & 56.6   & 37.6   & 77.9   & 50.7  & 31.7 \\
    \textit{Deep0Tag}~\cite{rahman2019deep0tag} & 34.4   & 58.7   & 43.4  & 22.6  & 64.0  & 33.4  & - \\
    \textit{LESA} (M = 1)~\cite{huynh2020shared}                        & 51.4   & 63.9   & 57.0   & 37.9   & 78.6   & 51.2  & 29.6 \\
    \textit{LESA} (M = 10)~\cite{huynh2020shared}                       & 52.3   & 65.1   & 58.0   & 38.6   & 80.0   & 52.0  & 31.5 \\
    \textit{BiAM}~\cite{narayan2021discriminative}                       & -   & -   & 59.6   & -   & -   & 53.4  & 47.8 \\
    \bf{Ours}                                & \bf{54.4}& \bf{67.1}& \bf{60.1}     & \bf{39.8} & \bf{82.5}& \bf{53.7}     & \bf{48.4} \\
    \bottomrule
    \end{tabular}}%
      \vspace{-5px}
  \label{tab4}%
\end{table*}
\subsection{Multi-Label Classification}
As a prediction model, we need to accurately classify seen classes to realize the prediction of unseen classes. Therefore, the multi-label classification ability of the model is also important. Table~\ref{tab4} shows the performance of our model under multi-label learning. For comparison, we added some classic multi-label classification algorithms, including \textit{Logistic Regression}~\cite{tsoumakas2007multi}, \textit{WSABIE}~\cite{weston2011wsabie}, \textit{WARP}~\cite{gong2013deep} and \textit{CNN-RNN}~\cite{wang2016cnn}, and some MLZSL models. We can see from the experimental results that our method has achieved the best results compared with the traditional multi-label learning model in both F1-Score and mAP. Compared with \textit{BiAM}~\cite{narayan2021discriminative}, the current best-performing method in MLZSL, our model still has considerable advantages. This is because, compared to \textit{BiAM}~\cite{narayan2021discriminative}, local and global features are not effectively fused, but through simple concatenation, we not only generate more effective semantic information, but also establish associations for them.

\subsection{Qualitative Performance}
\noindent \textbf{Attention Visualization: }Figure~\ref{fig:attmap} shows the performance of our model on attention visualization. First look at Figure~\ref{fig:attmap} (a), the response intervals of `animal' and `birds' have a large overlap, but `birds' pays more attention to the whole of the object, such as wings. This shows that there are both correlations between labels and their own unique expressions. Meanwhile, the corresponding regions of `clouds' and `sky' are also roughly close, confirming the effectiveness of our method. And Figure~\ref{fig:attmap} (b) shows the accuracy of our method on the corresponding region. Figure~\ref{fig:attmap} (c) is a special case where there are a large number of unseen labels. In this situation, the generated attention area and the corresponding label can still guarantee a complete fit. 

\noindent \textbf{Qualitative Results: } Our model's qualitative performance are shown in Figure~\ref{fig:quali}. For the input image, the label output by the model includes not only seen labels, but also unseen labels. As can be seen from the first image, the model first recognized `birds', and then other classes related to its semantics were also easily recognized. In the second image, from the point of view of the label, our model can not only recognize the class of specific mountains, such as `volcano', but also recognize very global information, such as the color `blue'. The third image accurately identifies the gender of the character. The success of these recognitions undoubtedly reflects the effectiveness of our combination of global and local semantics.

\begin{figure}[htbp]
    \centering
    \includegraphics[width=.48\textwidth]{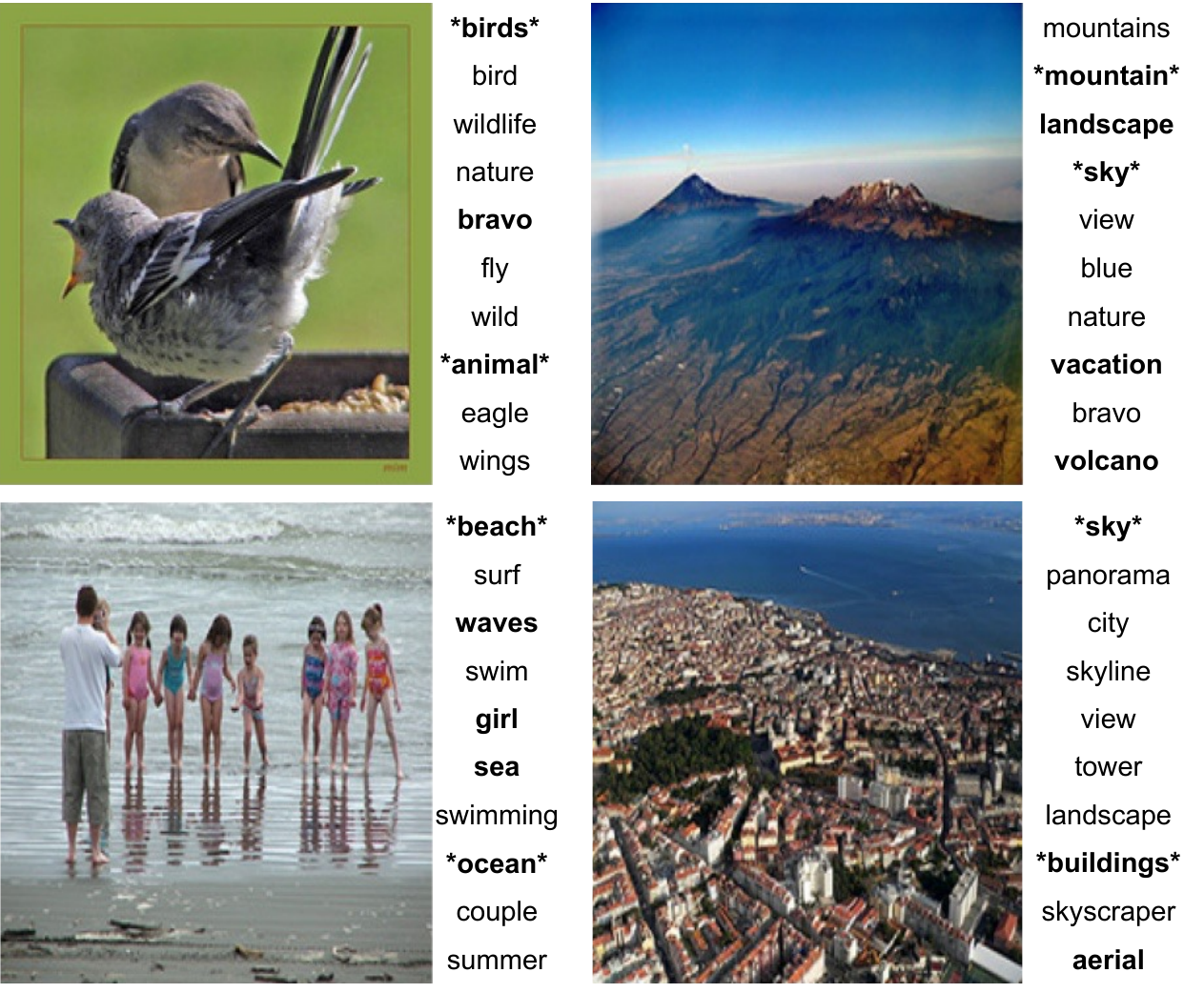}
    \caption{\textbf{Qualitative results.} The top-10 labels predicted by GBE-MLZSL in the case of Generalized MLZSL are shown above. The asterisk marks indicate unseen labels, while the bold ones indicate successfully predicted seen and unseen labels.}
    \vspace{-5px}
    \label{fig:quali}
\end{figure}

\section{Conclusion}
In this paper, we designed GBE-MLZSL to solve the problem of only relying on the principal direction of feature vectors for prediction and not being able to efficiently fuse local information with global information in previous methods. In response to the above problems, we designed ML-FEF to enhance the quality of the obtained features, and then grouped the features through the LID Module to obtain split local semantic information and enhance semantic diversity. At the same time, the GE Module is proposed to strengthen the global features of the image, that is, the principal direction. Finally, the semantic information is effectively combined and established through the Global-Local Association Graph. On two publicly available large MLZSL datasets \textit{NUS-WIDE} and \textit{Open-Images}-V4, our experimental results far outperform other state-of-the-art models. 


\bibliographystyle{IEEEtran}
\bibliography{thisbib.bib}

\end{document}